\documentclass{article}

\usepackage[preprint]{neurips_2026}
\usepackage{graphicx}
\usepackage{booktabs}

\usepackage{microtype}
\usepackage{graphicx}
\usepackage{subcaption}
\usepackage{multirow}

\usepackage{amsmath,amssymb}
\usepackage{bm}
\usepackage{enumitem}
\usepackage{xcolor}

\usepackage[utf8]{inputenc} % allow utf-8 input
\usepackage[T1]{fontenc}    % use 8-bit T1 fonts
\usepackage{hyperref}       % hyperlinks
\usepackage{url}            % simple URL typesetting
\usepackage{booktabs}       % professional-quality tables
\usepackage{amsfonts}       % blackboard math symbols
\usepackage{nicefrac}       % compact symbols for 1/2, etc.
\usepackage{microtype}      % microtypography
\usepackage{xcolor}         % colors

\title{SteeringDiffusion: A Bottlenecked Activation Control Interface for Diffusion Models}

\author{Fangzheng Wu \\
Department of Computer Science \\
Tulane University \\
\texttt{fwu6@tulane.edu} \\
\And
Brian Summa \\
Department of Computer Science \\
Tulane University \\
\texttt{bsumma@tulane.edu}
}

\begin{document}

\maketitle

\begin{abstract}
We introduce SteeringDiffusion, a bottlenecked activation-level control interface for diffusion models that exposes a smooth, monotonic, and runtime-adjustable control surface over the content--style trade-off.
Our method keeps the U-Net backbone frozen and learns a small, prompt-conditioned latent code projected to FiLM/AdaGN-style modulation parameters. A zero-initialized design guarantees exact equivalence to the base model at zero scale, while timestep-aware gating restricts modulation to later denoising stages. A single scalar at inference continuously traverses the control surface without retraining.
Across experiments on Stable Diffusion~1.5 and SDXL covering multiple artistic styles, we show that SteeringDiffusion produces smooth and monotonic content--style trade-offs. Under matched parameter budgets, it outperforms LoRA in controllability and stability, while ControlNet and rank-1 adapters do not expose a comparable control surface. We further introduce an inversion-stability diagnostic based on DDIM inversion, used as a post-hoc trajectory probe, which reveals strong correlations with intervention magnitude.
These results position \emph{Steering Bottlenecked Explicit Control (S-BEC)} as a practical, general-purpose control interface for frozen diffusion backbones.
\end{abstract}

% =====================================================================
% 1. INTRODUCTION
% =====================================================================
\section{Introduction}
\label{sec:introduction}

\begin{figure}[t]
  \centering
  \includegraphics[width=0.6\linewidth]{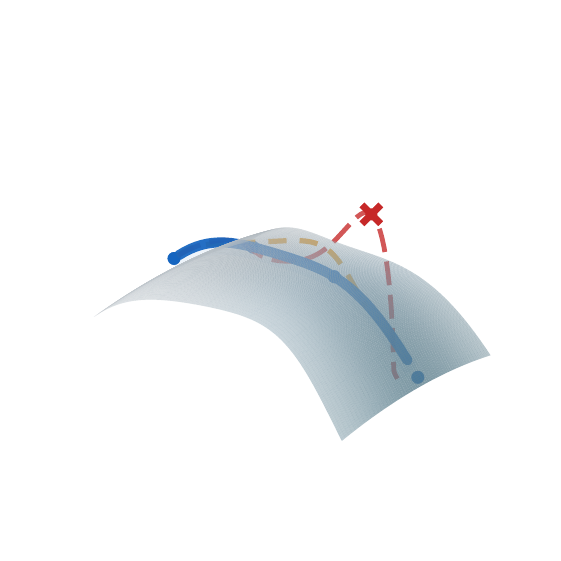}
  \vspace{-0.8em}
  \caption{\textbf{SteeringDiffusion exposes a smooth runtime control manifold.}
  \emph{Conceptual illustration (not to scale).} Varying a single inference-time scalar traverses a continuous,
  monotonic content--style trade-off (blue), while alternative control mechanisms can exhibit fragmented traversal
  (red) or off-manifold, more invasive behavior (orange). Quantitative trade-off curves and matched-strength
  comparisons are provided in Fig.~\ref{fig:pareto_lora} and the Supplementary Material.}
  \label{fig:teaser}
  \vspace{-1.0em}
\end{figure}

Diffusion models~\cite{ho2020ddpm,song2021score,dhariwal2021diffusion,rombach2022ldm,karras2022edm} have become the dominant paradigm for high-quality image synthesis.
However, adapting these models to new visual styles or domains remains challenging.
Full fine-tuning is computationally expensive and risks overfitting, while popular parameter-efficient fine-tuning (PEFT) methods---such as LoRA~\cite{hu2022lora}, DiffFit~\cite{mou2023diffit}, or orthogonal fine-tuning~\cite{qiu2023oft}---do not expose a continuous, traversable control surface at inference time: changing the operating point requires retraining or coarse hyperparameter tuning rather than moving along a coherent control axis.
Other approaches such as DreamBooth~\cite{ruiz2023dreambooth}, Textual Inversion~\cite{gal2022textualinversion}, Custom Diffusion~\cite{kumari2023custom}, and SVDiff~\cite{han2023svdiff} similarly lack an explicit control interface that allows continuous traversal of content--style trade-offs without modifying model parameters.
Fig.~\ref{fig:teaser} illustrates our goal: a \emph{safe-to-disable} runtime control interface that yields a smooth,
monotonic content--style trade-off while remaining exactly equivalent to the base model at zero scale.

% --- Teaser (Figure 1) ---

\subsection*{Activation-level steering as an alternative}

In contrast, recent work in large language models has shown that direct interventions on internal activations---rather than weights---can expose and control latent behaviors of a frozen model~\cite{turner2023activation,sinii2025steering,zou2023representation,li2024iti,subramani2022extracting}.
So-called \emph{activation steering} methods demonstrate that adding or modulating hidden representations along learned directions can induce systematic behavioral changes without modifying the underlying parameters~\cite{rimsky2024steering,todd2024function,hernandez2024linearity,park2024linear}.
This suggests that a low-dimensional activation interface, rather than weight updates, may be the appropriate abstraction for controllable generative models.

Motivated by this perspective, we ask:
\begin{quote}
\emph{Can diffusion models be steered through activation-level modulation to expose a smooth, monotonic control surface?}
\end{quote}

\subsection*{Steering diffusion via activation-space modulation}

We formulate diffusion controllability through the lens of Steering via Bottlenecked Explicit Control (S-BEC), defining a control-interface abstraction for frozen diffusion models.
Rather than modifying U-Net weights, our approach introduces a small, prompt-conditioned latent code that modulates intermediate activations via FiLM/AdaGN-style affine transformations~\cite{perez2018film,huang2017adain}.
The diffusion backbone remains entirely frozen.

Three design principles guide our approach:
\begin{itemize}
    \item \textbf{Non-invasiveness.}
    The steering mechanism is zero-initialized and preserves exact equivalence to the frozen model at initialization (E0). No pretrained weights are altered.

    \item \textbf{Low-dimensional control.}
    Steering is mediated by a compact latent code $\mathbf{v} \in \mathbb{R}^k$.
    Empirically, increasing $k$ beyond $8$--$16$ yields no statistically significant gains, suggesting that a small shared code suffices for effective steering.

    \item \textbf{Temporal awareness.}
    Steering is gated by diffusion timestep, emphasizing late denoising stages where appearance and texture are most malleable.
\end{itemize}

The strength of steering is controlled by a single scalar $s$ at inference time, enabling a continuous and interpretable trade-off between content preservation and style strength without retraining.

\subsection*{Empirical findings}

We conduct a systematic empirical study on Stable Diffusion~1.5 using ArtBench~\cite{liao2022artbench}, with generalization to SDXL.
Our results show that:
(i)~activation-level steering exposes a smooth and monotonic runtime trade-off curve;
(ii)~under matched parameter budgets, steering is competitive with or superior to LoRA, with 30--36\% higher style shift at comparable content preservation;
(iii)~ControlNet~\cite{zhang2023controlnet} and rank-1 adapters~\cite{lyu2024onedim} fail to expose a comparable control surface; and
(iv)~an inversion-stability diagnostic based on DDIM inversion~\cite{song2021ddim} reveals trajectory-level properties complementary to perceptual metrics.

\subsection*{Contributions}

\begin{itemize}
    \item We study \textbf{activation-level steering for diffusion models} through the lens of Steering via Bottlenecked Explicit Control (S-BEC), demonstrating that frozen diffusion backbones can be controlled through lightweight activation modulation.

    \item We provide a \textbf{systematic analysis of the control surface}, showing smooth, monotonic, and style-invariant trade-off behavior.

    \item We introduce an \textbf{inversion-stability diagnostic} that characterizes trajectory perturbation complementary to perceptual metrics.

    \item We demonstrate that \textbf{existing methods (LoRA, ControlNet, rank-1) cannot expose a comparable control surface}, positioning S-BEC as a distinct control paradigm.
\end{itemize}

% =====================================================================
% 2. RELATED WORK (compressed)
% =====================================================================
\section{Related Work}
\label{sec:related_work}

\subsection{Parameter-Efficient Adaptation and Activation Steering}

We view parameter-efficient adaptation and control of generative models through
a unified residual perspective.
Let $h_\ell$ denote the hidden representation at layer $\ell$ and
$\mathcal{F}_\ell(h_\ell; \theta_\ell, c)$ the frozen backbone update conditioned on
text $c$. Many adaptation approaches can be written as
\begin{equation}
    h_{\ell+1} = h_\ell + \mathcal{F}_\ell(h_\ell; \theta_\ell, c) + \Delta_\ell(h_\ell, c, t),
    \label{eq:residual_steering}
\end{equation}
where $\Delta_\ell$ is a task-specific steering term with a small parameter
footprint.

\begin{figure}[t]
  \centering
  \includegraphics[width=0.9\linewidth]{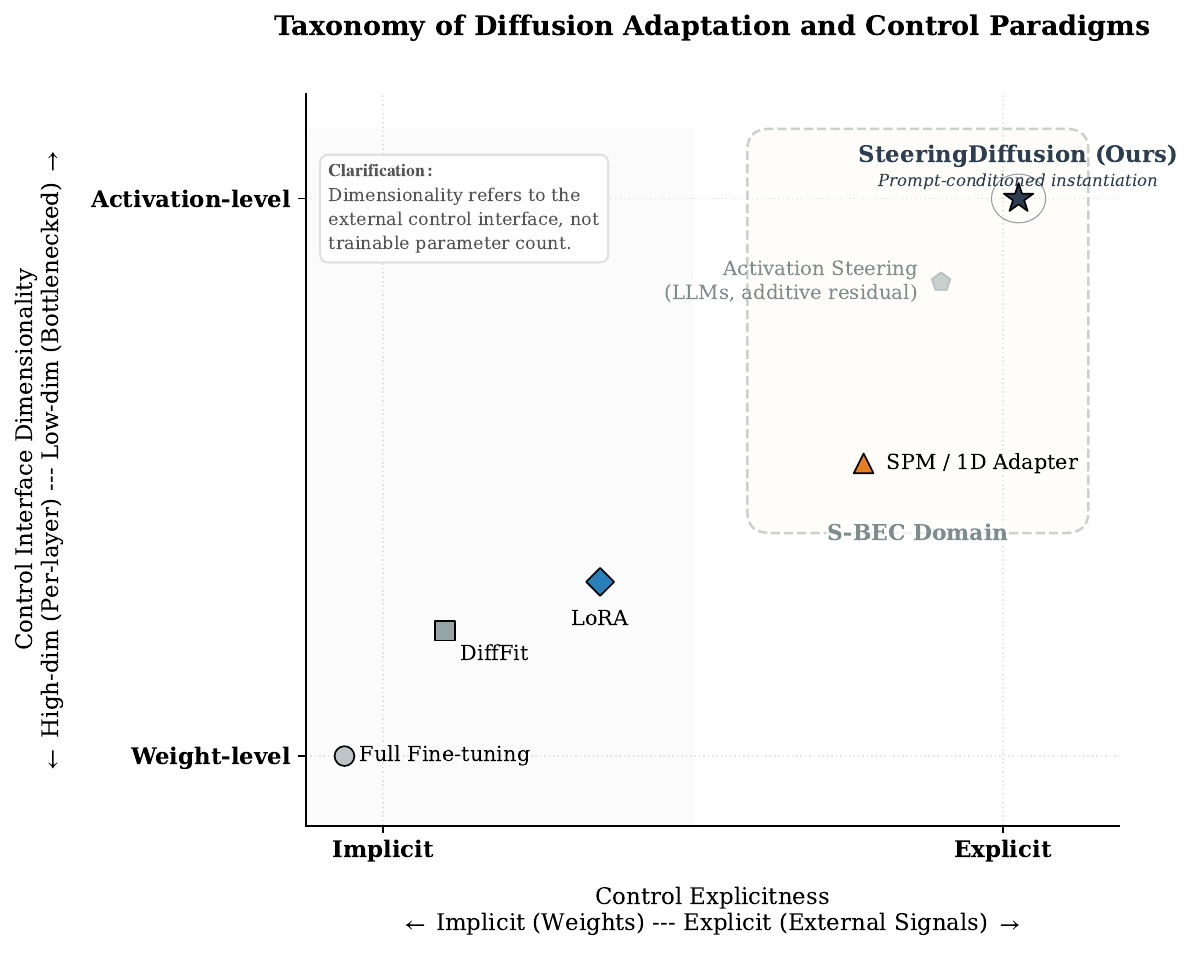}
  \caption{
    \textbf{Taxonomy of parameter-efficient adaptation and steering mechanisms.}
    Methods are organized by whether control is implicit in weight space or explicit
    via external signals (x-axis), and by the dimensionality of the control bottleneck
    (y-axis). SteeringDiffusion belongs to bottlenecked explicit control (S-BEC).
  }
  \label{fig:sbectaxonomy}
\end{figure}

\paragraph{Steering via Bottlenecked Explicit Control (S-BEC).}
A growing line of work can be interpreted through the lens of
\emph{steering via bottlenecked explicit control} (S-BEC).
Methods in this family steer a frozen backbone using a low-dimensional control
signal whose strength can be adjusted externally at inference time.
In language models, activation steering~\cite{turner2023activation,sinii2025steering},
representation engineering~\cite{zou2023representation},
inference-time intervention~\cite{li2024iti},
contrastive activation addition~\cite{rimsky2024steering},
and latent steering vectors~\cite{subramani2022extracting}
all demonstrate that low-dimensional residual-stream edits can elicit controlled behavioral changes.
The linear representation hypothesis~\cite{park2024linear,hernandez2024linearity,todd2024function}
provides theoretical grounding for why such low-dimensional interventions are effective.
For diffusion models, one-dimensional adapters~\cite{lyu2024onedim}
and concept erasure methods~\cite{gandikota2023erasing,gandikota2024unified}
represent early instances of explicit low-dimensional control.
Figure~\ref{fig:sbectaxonomy} summarizes this taxonomy.

\paragraph{LoRA and weight-space adaptation.}
Weight-space PEFT for diffusion includes low-rank updates
(LoRA~\cite{hu2022lora}), compact SVD-structured parameterizations
(SVDiff~\cite{han2023svdiff}), orthogonal fine-tuning~\cite{qiu2023oft},
Kronecker-structured updates~\cite{marjit2024diffusekrona},
and subject-driven personalization~\cite{ruiz2023dreambooth,kumari2023custom,gal2022textualinversion}.
These methods provide flexible domain adaptation but typically learn
high-dimensional, per-layer updates tied to individual weight matrices
and require retraining per operating point.
In contrast, our approach factorizes steering through a single $k$-dimensional
code shared across all steered blocks, exposing an external runtime control variable.

\paragraph{One-dimensional adapters and concept control.}
SPM~\cite{lyu2024onedim} attaches a rank-1 direction
to each layer, implementing $\Delta_\ell(h_\ell) = \alpha_\ell \cdot s_\ell$.
Related concept erasure methods~\cite{gandikota2023erasing,gandikota2024unified},
building on insights from network dissection~\cite{bau2017network} and knowledge editing~\cite{meng2022rome},
modify model internals to suppress specific concepts.
These methods target concept erasure rather than continuous style injection,
and their control is typically fixed-direction and not prompt-conditioned.

\paragraph{Diffusion controllability and editing.}
A parallel line of work seeks to control diffusion outputs through
guidance~\cite{ho2022classifierfree,dhariwal2021diffusion},
attention manipulation~\cite{hertz2022prompt,aberman2023pnp},
image-conditioned adapters~\cite{ye2023ipadapter,mou2024t2iadapter,zhang2023controlnet},
or text-driven editing~\cite{brooks2023instructpix2pix,kawar2023imagic,meng2022sdedit}.
Diffusion-based style transfer methods~\cite{wang2023stylediffusion,wang2022diffusionclip,patashnik2021styleclip}
focus on content--style disentanglement or training-free controllability.
Our focus is orthogonal: rather than editing individual images, we expose a
monotonic runtime-adjustable control surface via an explicit bottlenecked
activation interface.

\paragraph{Activation steering in language models.}
Recent work~\cite{sinii2025steering,zou2023representation,li2024iti} shows that adding learned vectors to the residual stream of frozen transformers
can elicit new behaviors with extremely small parameter budgets.
Our method adapts this perspective to diffusion models, where continuous denoising trajectories replace discrete token sequences.

\paragraph{Our steering mechanism.}
Within~\eqref{eq:residual_steering}, our method implements a prompt- and
timestep-dependent activation adapter.
A pooled text embedding is mapped to a low-dimensional steering code
$v = g_\theta(c) \in \mathbb{R}^k$, which is projected to FiLM/AdaGN-style affine
parameters~\cite{perez2018film,huang2017adain,park2019spade}:
\begin{equation}
    \Delta_\ell(h_\ell, c, t) =
    s\, f(t)\bigl(\gamma_\ell(v) \odot \mathrm{GN}(h_\ell) + \beta_\ell(v)\bigr),
    \label{eq:steering_delta}
\end{equation}
where $f(t)$ gates modulation toward late denoising stages and $s$ is an
inference-time scale.
This design (i)~uses a shared low-dimensional control variable,
(ii)~is explicitly timestep-aware, and
(iii)~exposes a smooth content--style trade-off traversable at inference
time without retraining.

% =====================================================================
% 3. EXPERIMENTS
% =====================================================================
\section{Style Steering Experiments on SD~1.5}
\label{sec:exp-sd15-steering}

\subsection{Experimental Setup}
\label{sec:exp-setup}

\paragraph{Backbone and steering adapter.}
We build on Stable Diffusion v1.5~\cite{rombach2022ldm}, using the public UNet, VAE and CLIP~\cite{clip} text encoder as a frozen backbone.
On top of the frozen UNet we add a lightweight \emph{SteeringInjector} module with two components:
(i)~a per-prompt code generator $g_\theta$,
and (ii)~per-layer AdaGN-style steering projections.

Given a text prompt, we compute token embeddings via the frozen CLIP text encoder and average pool them to obtain a global condition vector
$c \in \mathbb{R}^{d_\text{text}}$.
The code generator $g_\theta$ is a two-layer MLP with SiLU nonlinearity,
mapping $c$ to a $k$-dimensional steering code $v \in \mathbb{R}^k$.
We attach steering modules to middle and up-sampling blocks (\texttt{mid\_up}).
Each attached block applies AdaGN-style steering to the GroupNorm-normalized features:
\begin{equation}
    \Delta = s \, f(t) \left( \gamma(v) \odot \tilde{h} + \beta(v) \right),
    \label{eq:steering_residual}
\end{equation}
where $\tilde{h} = \mathrm{GN}(h)$, and $\gamma, \beta$ are zero-initialized linear projections ensuring exact equivalence to the base model at zero steering scale.
Only $g_\theta$ and the steering projections are trainable.

\paragraph{Timestep-aware gating.}
A fixed schedule $f(t) \in [0,1]$ (shifted-scaled sigmoid) suppresses steering at early timesteps and ramps it up near the end of the trajectory.
Setting $s=0$ recovers the frozen baseline exactly.

\paragraph{Dataset.}
We focus on ArtBench-10~\cite{liao2022artbench} style classes, with all images resized to $512\times512$.
We follow the standard DDPM~\cite{ho2020ddpm} training objective on the frozen VAE latent space.
Full training details (optimization, hyperparameters, DDIM/CFG settings) are provided in Supplementary Sec.~A.

% -----------------------------------------------------------------
\subsection{Evaluation Metrics}
\label{sec:metrics}

We evaluate steering quality using CLIP-based metrics computed with ViT-L/14~\cite{clip}.

\paragraph{Content preservation (CLIP-I).}
Cosine similarity between CLIP image embeddings of steered and baseline outputs:
\begin{equation}
    \text{CLIP-I} = \cos\bigl(f(x_{\text{steer}}),\, f(x_{\text{base}})\bigr).
    \label{eq:clip-i}
\end{equation}

\paragraph{Text alignment (CLIP-T).}
Cosine similarity between the steered image and the text prompt:
\begin{equation}
    \text{CLIP-T} = \cos\bigl(f(x_{\text{steer}}),\, g(\text{prompt})\bigr).
    \label{eq:clip-t}
\end{equation}

\paragraph{Style shift.}
We construct a style prototype $\mu_{\text{style}}$ by averaging CLIP image embeddings over held-out style reference images.
Style shift measures the positive increase in similarity to this prototype relative to the baseline:
\begin{equation}
    \Delta_{\text{style}} = \max\Bigl(0,\;
        \cos\bigl(f(x_{\text{steer}}), \mu_{\text{style}}\bigr)
        - \cos\bigl(f(x_{\text{base}}), \mu_{\text{style}}\bigr)
    \Bigr).
    \label{eq:style-shift}
\end{equation}

\paragraph{Inversion-stability diagnostic (Inv-Stab).}
As a post-hoc trajectory diagnostic---not a quality metric---we measure how much steering perturbs the denoising trajectory.
Adapted from~\cite{zhang2025notallnoises}, given the initial latent $z_T$ sampled at inference time and its recovered estimate $\hat{z}_T$ obtained by DDIM inversion~\cite{song2021ddim,mokady2022nulltext} of the generated image:
\begin{equation}
\text{Inv-Stab} = \cos(z_T, \hat{z}_T).
\label{eq:inv-stab}
\end{equation}
This is never used as a training signal. We validate its diagnostic properties in Supplementary Sec.~D.
Prototype robustness (varying the number of reference images and random subsets) and LPIPS consistency are validated in Supplementary Sec.~E.1, showing that the observed trade-offs are not artifacts of CLIP embedding choice or prototype construction.

% -----------------------------------------------------------------
\subsection{Exact Zero-Scale Equivalence and Parameter Count}
\label{sec:zero-equiv-param}

We zero-initialize all steering projections and verify that the maximum absolute
difference between the frozen UNet and UNet+SteeringInjector outputs is numerically
zero (up to floating point round-off), confirming exact equivalence to the base
model at zero steering scale.

\begin{table}[t]
  \centering
  \small
  \begin{tabular}{lrr}
    \toprule
    Configuration & Trainable params & \% of UNet \\
    \midrule
    Full SD~1.5 UNet (frozen) & $859.5$M & $100\%$ \\
    Steering-B, $k{=}64$, mid\_up &
      $2.92$M & $0.34\%$ \\
    \bottomrule
  \end{tabular}
  \caption{Trainable parameter counts. Only the steering parameters are trained; the UNet, VAE, and text encoder are frozen.}
  \label{tab:sd15-param-count}
\end{table}

% -----------------------------------------------------------------
\subsection{Control Surface: Smooth Monotonic Trade-off}
\label{sec:control-surface}

The central claim of this paper is that activation-level steering exposes a smooth, monotonic control surface.
Figure~\ref{fig:pareto_lora} (left) directly demonstrates this property: as the steering scale $s$ increases, CLIP-I decreases monotonically while Style Shift increases monotonically, with no abrupt transitions or saturation effects.
CLIP-T remains stable across the entire range.

This behavior is consistent across steering dimensions ($k \in \{8, 16, 32, 64\}$) and styles.
We use \texttt{mid\_up} as a conservative default; layer-wise ablations across five configurations show consistent trade-off geometry with CLIP-I variance below 0.02 (Supplementary Sec.~B.1).
Full scale sweep and $k$-sweep tables confirm that increasing $k$ beyond 16 yields no statistically significant gains (paired Wilcoxon signed-rank test, $p > 0.3$), validating that a small shared code suffices (Supplementary Sec.~B.2).

% -----------------------------------------------------------------
\subsection{Comparison with LoRA}
\label{sec:lora-comparison}

We compare against LoRA~\cite{hu2022lora} as a representative weight-space PEFT baseline.
Both methods are trained on the same Art Nouveau subset of ArtBench-10
with identical training budgets (5k steps, batch size 4, lr $5\times10^{-5}$).
Steering uses $k{=}16$ (0.88M params); LoRA uses $r{=}4$ (0.80M params).

\begin{figure}[t]
\centering
\includegraphics[width=\linewidth]{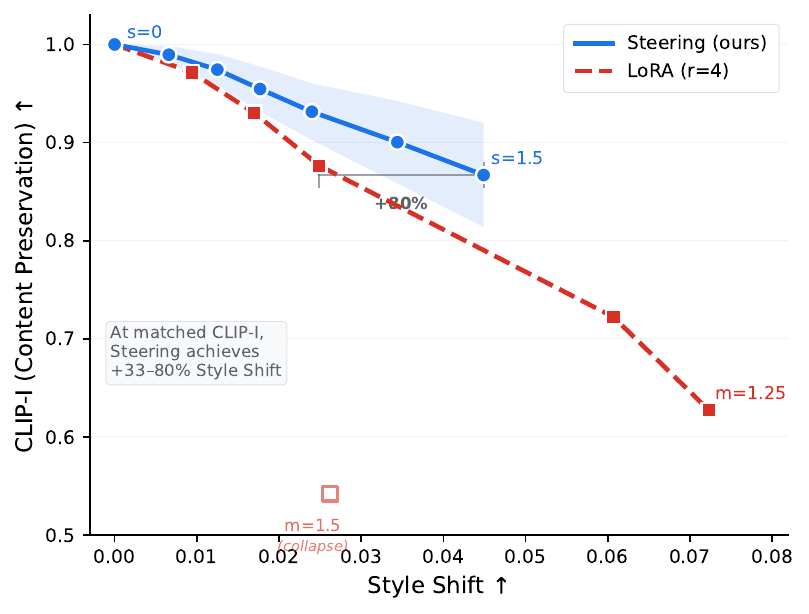}
\caption{Content--style trade-off on Art Nouveau (class~0).
At matched CLIP-I, steering achieves 33--80\% higher Style Shift than LoRA ($r=4$).
LoRA exhibits non-monotonic behavior at high multipliers ($m\ge1.0$).}
\label{fig:pareto_lora}
\end{figure}

\paragraph{Results.}
Figure~\ref{fig:pareto_lora} shows the content--style trade-off curves.
At matched content preservation (CLIP-I),
steering consistently achieves higher style transfer.
Table~\ref{tab:matched_comparison} reports three matched regimes
where steering yields 33--80\% higher Style Shift.

Steering exhibits a smooth, monotonic trade-off as $s$ increases.
By contrast, LoRA shows a sharp drop in content preservation between
$m=0.75$ and $m=1.0$, and enters a failure regime at $m=1.5$
where Style Shift becomes non-monotonic---a behavior we term \emph{style collapse}.

\begin{table}[t]
\centering
\caption{Matched operating point comparison (by CLIP-I).
Steering consistently achieves stronger style transfer.}
\label{tab:matched_comparison}
\small
\begin{tabular}{@{}ccccccc@{}}
\toprule
\multicolumn{3}{c}{CLIP-I} & \multicolumn{2}{c}{Style Shift} & \multirow{2}{*}{$\Delta$} & \multirow{2}{*}{Gain} \\
\cmidrule(lr){1-3}\cmidrule(lr){4-5}
Steering & LoRA & $|\Delta|$ & Steering & LoRA & & \\
\midrule
0.974 ($s{=}0.5$) & 0.972 ($m{=}0.25$) & 0.003 & 0.0125 & 0.0094 & +0.003 & +33\% \\
0.931 ($s{=}1.0$) & 0.930 ($m{=}0.5$)  & 0.001 & 0.0240 & 0.0170 & +0.007 & +41\% \\
0.867 ($s{=}1.5$) & 0.877 ($m{=}0.75$) & 0.0096 & 0.0449 & 0.0249 & +0.020 & +80\% \\
\bottomrule
\end{tabular}
\end{table}

Beyond the $r{=}4$ setting reported here, Supplementary Sec.~C.1 evaluates LoRA across ranks $r \in \{2, 4, 8\}$, multipliers up to $m{=}2.0$, two CLIP backbones (ViT-B/32, ViT-L/14), and multiple layer placements.
In all configurations, LoRA exhibits either weak stylization at low $m$ or non-monotonic collapse at higher $m$, whereas SteeringDiffusion maintains a smooth trade-off throughout.
Supplementary Sec.~C.1 further evaluates rsLoRA~\cite{kalajdzievski2023rslora}, which uses rank-stabilized scaling ($\alpha/\sqrt{r}$ instead of $\alpha/r$); the non-monotonic behavior persists, confirming that the control surface limitation is inherent to weight-space adaptation rather than a scaling artifact.

\paragraph{rsLoRA with matched effective scaling.}
To address the recently proposed rank-stabilized LoRA
(rsLoRA)~\cite{kalajdzievski2023rslora}, which replaces
$\alpha/r$ with $\alpha/\sqrt{r}$, we additionally evaluate rsLoRA
under a matched-effective-scaling protocol
($\mathrm{eff} = m \cdot \alpha/\sqrt{r}$)
that aligns the effective perturbation strength point-by-point with
standard LoRA.
Even under this matched protocol, rsLoRA does not produce a
consistently monotonic control surface: Style Shift increases through
$\mathrm{eff}{=}1.25$ but reverses at $\mathrm{eff}{=}1.50$
($0.035 \to 0.027$), while structural degradation continues to
grow (LPIPS $0.377 \to 0.489$).
Full results are in Supplementary Material.

% -----------------------------------------------------------------
\subsection{ControlNet and Rank-1 Baselines}
\label{sec:baselines}

\paragraph{ControlNet baseline.}
A ControlNet-style branch~\cite{zhang2023controlnet} trained on the same data requires 720$\times$ more parameters and exhibits substantially larger content disruption at equivalent style shifts.
Supplementary Sec.~C.3 further evaluates ControlNet on all 553 prompts, showing that inversion stability degrades sharply with increasing style strength even when CLIP-I is matched, confirming that the branch-based architecture disrupts denoising trajectories far more than activation-level steering.

\paragraph{Rank-1 feature adapter.}
A minimal rank-1 direction extracted from steering residuals achieves only $\sim$69\% of our style shift at matched CLIP-I.
Supplementary Sec.~C.2 reports a stronger rank-1 baseline with large-scale $\alpha$ sweeps ($\alpha \in \{100, 300, 500\}$) and stability analysis, confirming that one-dimensional adapters achieve noticeable style only at the cost of severe trajectory disruption (Inv-Stab drops below 0.80).

\begin{table}[t]
\centering
\caption{Comparison with ControlNet and rank-1 baselines. Steering achieves superior style transfer with dramatically lower parameter count.}
\label{tab:baseline_summary}
\small
\begin{tabular}{@{}lccc@{}}
\toprule
Method & Params & CLIP-I & Style Shift \\
\midrule
ControlNet & 361M & 0.59 & 0.045 \\
Rank-1 ($\alpha{=}500$) & 0 & 0.849 & 0.031 \\
Ours ($s$ matched) & 0.5M & 0.850 & \textbf{0.045} \\
\bottomrule
\end{tabular}
\end{table}

These results validate that effective style control requires a \emph{bottlenecked, condition-aware} interface rather than heavy external branches or fixed directional biases.

\paragraph{Training-free skip-feature injection.}
We also compare against a representative training-free baseline
that blends UNet skip connections from a style-reference trajectory
into the content trajectory~\cite{aberman2023pnp}.
At matched content preservation (CLIP-I ${\approx}0.88$),
skip injection achieves comparable Style Shift ($0.022$ vs.\ $0.025$)
but with $2.9{\times}$ higher structural degradation
(LPIPS $0.40$ vs.\ $0.14$) and $45{\times}$ greater inference cost
($2.2$\,s vs.\ $49$\,ms per image).
Full analysis is in Supplementary Material.

% -----------------------------------------------------------------
\subsection{Multi-Style Generalization}
\label{sec:multistyle}

A natural question: is the smooth control surface specific to one style, or does it generalize?
We evaluate steering on four visually distinct ArtBench styles---Art Nouveau (class~0),
Impressionism (class~3), Renaissance (class~6), and Ukiyo-e (class~9)---spanning
Western to Eastern, classical to modern.

\begin{figure}[t]
  \centering
  \includegraphics[width=0.95\linewidth]{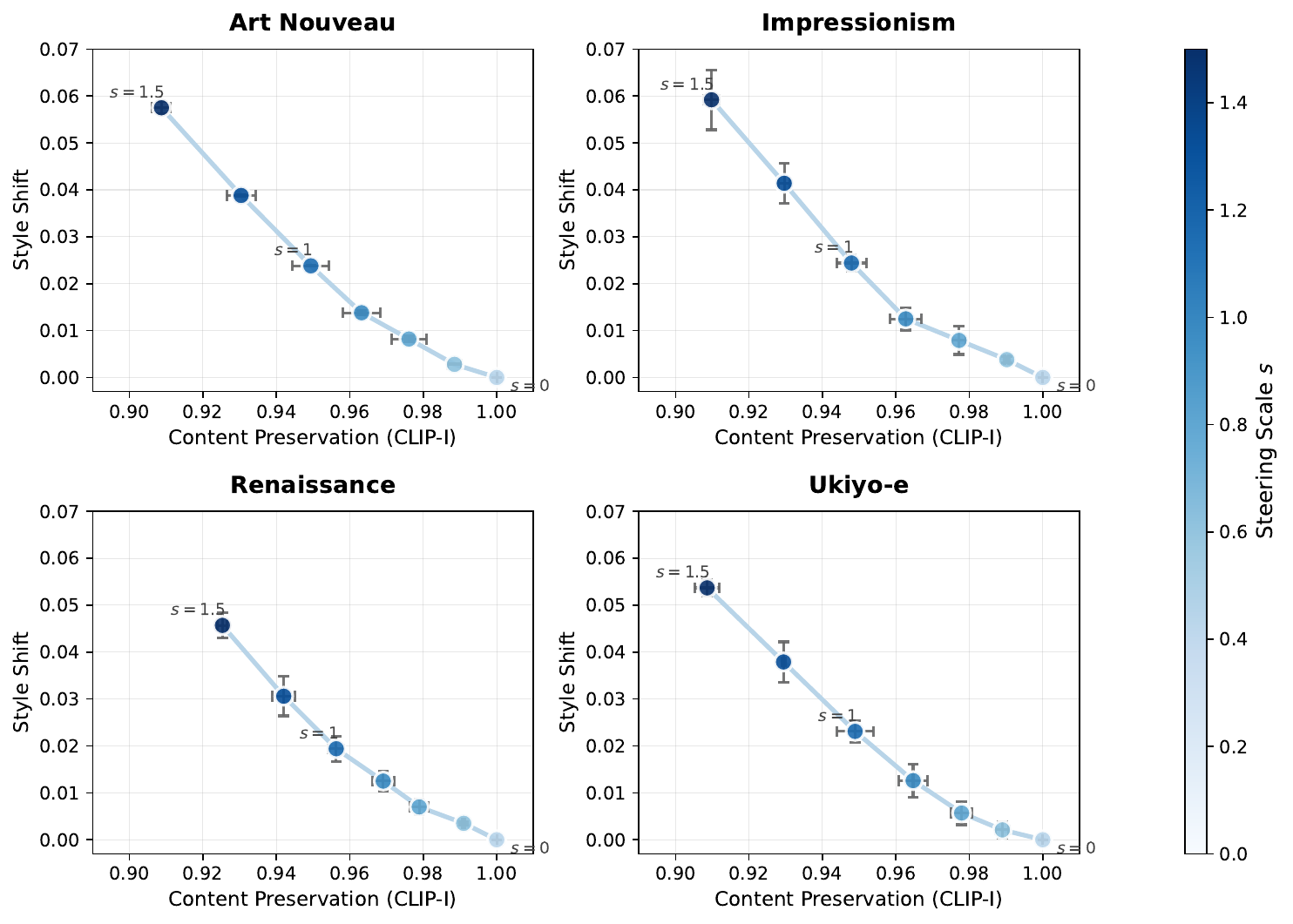}
  \caption{\textbf{Content--style trade-off across four ArtBench styles.}
  Increasing the steering scale $s$ consistently produces monotonic trade-offs
  across all styles, confirming a style-invariant control surface.}
  \label{fig:pareto_4style}
\end{figure}

\paragraph{Results.}
Figure~\ref{fig:pareto_4style} presents the content--style trade-off curves.
Across all styles, increasing $s$ leads to monotonic increases in Style Shift
and corresponding decreases in CLIP-I.
The effective operating regime ($s \approx 0.5$--$1.0$) transfers directly to all four styles.

\paragraph{Monotonicity analysis.}
Table~\ref{tab:monotonicity_4style} confirms strict monotonicity.
Denoting by $\rho$ the Spearman rank correlation,
$\rho(s, \text{Style Shift}) = +1.0$ and $\rho(s, \text{CLIP-I}) = -1.0$
with $p < 10^{-4}$ for all styles, and zero violations across all style--seed combinations.

\begin{table}[t]
  \centering
  \small
  \setlength{\tabcolsep}{6pt}
  \begin{tabular}{l c c c c}
    \toprule
    Style & $\rho(s, \text{Style})$ & $\rho(s, \text{CLIP-I})$ & Style Viol. & CLIP-I Viol. \\
    \midrule
    Art Nouveau     & $+1.000^{*}$ & $-1.000^{*}$ & 0/6 & 0/6 \\
    Impressionism   & $+1.000^{*}$ & $-1.000^{*}$ & 0/6 & 0/6 \\
    Renaissance     & $+1.000^{*}$ & $-1.000^{*}$ & 0/6 & 0/6 \\
    Ukiyo-e         & $+1.000^{*}$ & $-1.000^{*}$ & 0/6 & 0/6 \\
    \bottomrule
  \end{tabular}
  \caption{\textbf{Monotonicity statistics across four ArtBench styles.}
  $\rho$: Spearman rank correlation between scale $s$ and metric.
  $^{*}$Spearman test $p < 0.0001$. All styles exhibit strict monotonicity with zero violations.}
  \label{tab:monotonicity_4style}
\end{table}

Supplementary Sec.~B.3 reports full per-style scale sweeps and multiple random seeds, with no observed violations of monotonicity and cross-seed standard deviations below 0.01, confirming that the control surface is robust to both style choice and training stochasticity.

% -----------------------------------------------------------------
\subsection{Generalization to SDXL}
\label{sec:sdxl}

Table~\ref{tab:sdxl_results} confirms that the monotonic control surface generalizes to SDXL~\cite{podell2023sdxl} ($1024 \times 1024$, larger UNet) across two styles.

\begin{table}[t]
\centering
\caption{SDXL results. Monotonic controllability is preserved on a larger backbone.}
\label{tab:sdxl_results}
\small
\begin{tabular}{@{}llccc@{}}
\toprule
Style & Scale $\alpha$ & CLIP-T & CLIP-I & Style Shift \\
\midrule
\multirow{3}{*}{Art Nouveau}
 & 0.5 & 0.289 & 0.846 & 0.015 \\
 & 1.0 & 0.293 & 0.828 & 0.024 \\
 & 1.5 & 0.289 & 0.782 & 0.036 \\
\midrule
\multirow{3}{*}{Ukiyo-e}
 & 0.5 & 0.288 & 0.846 & 0.017 \\
 & 1.0 & 0.291 & 0.777 & 0.024 \\
 & 1.5 & 0.265 & 0.698 & 0.049 \\
\bottomrule
\end{tabular}
\end{table}

Supplementary Sec.~B provides full SDXL trade-off curves and qualitative galleries across additional scales and styles, confirming that the monotonic control surface persists on the larger backbone without architectural modifications.

% -----------------------------------------------------------------
\subsection{Runtime Efficiency}
\label{sec:runtime}

Our steering adapter ($k{=}32$, 1.56M params, $0.18\%$ of UNet) trains $\sim$2$\times$ faster than LoRA (91.7 vs.\ 125 ms/step on an RTX A6000) and adds only $\sim$3\,ms inference overhead over the frozen baseline (48.9 vs.\ 45.2\,ms). ControlNet requires $361$M parameters with proportionally higher costs.

% -----------------------------------------------------------------
\subsection{Interpreting Inversion Stability}
\label{sec:inv-interp}

Inv-Stab is used solely as a post-hoc trajectory diagnostic, not as a quality metric.
Across 553 prompts and seven steering scales, strong Spearman rank correlations with perceptual metrics emerge ($|\rho| > 0.65$), reflecting shared sensitivity to intervention magnitude; however, within a fixed scale, correlations vanish ($|\rho| < 0.12$), confirming that Inv-Stab captures complementary trajectory-level information rather than perceptual quality.
Full correlation matrices and per-scale breakdowns are in Supplementary Sec.~D.

\subsection{Discussion: Why Bottlenecked Activation Control Yields a Stable Control Surface}
\label{sec:discussion}

A central question raised by our results is \emph{why} bottlenecked activation-level steering produces a smooth and monotonic control surface, while weight-space and branch-based alternatives do not.
We discuss three complementary mechanisms.

\paragraph{Bottlenecked interfaces induce low-dimensional control geometry.}
In S-BEC, all steering decisions are mediated through a single low-dimensional latent code $v \in \mathbb{R}^k$, shared across all steered blocks.
This creates an explicit, global control interface whose dimensionality is independent of network depth or width.
In contrast, weight-space methods such as LoRA distribute control across thousands of independent parameters tied to individual layers and matrices.
Increasing the LoRA multiplier $m$ does not correspond to moving along a coherent direction in representation space, but rather to amplifying a heterogeneous set of layer-specific perturbations.
This explains the observed non-monotonicity and style collapse at high multipliers: the effective control manifold is high-dimensional and poorly aligned with perceptual axes.

\paragraph{Activation-level modulation preserves trajectory coherence.}
SteeringDiffusion operates by applying affine modulation to normalized activations along the denoising trajectory, rather than altering the parameters that define the trajectory itself.
This distinction is crucial.
Weight updates change the vector field of the diffusion dynamics, potentially introducing new attractors or unstable regions, whereas activation-level modulation perturbs the current state while keeping the underlying dynamics fixed.
ControlNet-style branches similarly inject large external signals that alter the effective trajectory in a nonlocal manner.
Our inversion-stability analysis (Sec.~\ref{sec:inv-interp}) supports this interpretation: both ControlNet and rank-1 adapters induce large trajectory deviations at comparable perceptual effects, whereas bottlenecked activation steering achieves control with substantially smaller trajectory perturbations.

\paragraph{Temporal gating aligns control with semantic emergence.}
The timestep-dependent gate $f(t)$ plays a key role in stabilizing the control surface.
Early denoising steps primarily establish global structure and object layout, while later steps refine texture and style~\cite{zhang2025notallnoises}.
By suppressing steering at early timesteps and concentrating modulation near the end of the trajectory, S-BEC aligns its control with stages where appearance is most malleable and least disruptive to content.
This temporal separation explains why SteeringDiffusion maintains monotonic trade-offs across styles and backbones, while untimed or uniformly applied perturbations are prone to entanglement and instability.

\medskip
Taken together, these factors suggest that controllability in diffusion models is fundamentally an \emph{interface design} problem rather than a parameter-count problem.
A small but explicit, bottlenecked, and temporally aligned control interface can expose a well-behaved control surface even on a large frozen backbone, whereas high-capacity but implicit control mechanisms fail to provide reliable or interpretable traversal of the content--style trade-off.

% -----------------------------------------------------------------
\subsection{Limitations and Scope}
\label{sec:limitations}

We explicitly delineate the claims and boundaries of this work to avoid overclaiming generality.

\paragraph{Style-specific validation.}
All quantitative evaluations use artistic style transfer as the control target.
While the S-BEC framework is defined generally---any attribute expressible as a directional shift in activation space could in principle be steered---we do \emph{not} claim that the smooth monotonic control surface demonstrated here transfers automatically to arbitrary control targets such as object identity, spatial layout, or fine-grained attribute editing.
Each new target would require its own empirical validation.

\paragraph{Single-attribute steering.}
Our current formulation steers a single attribute (style) via one shared latent code $v$.
Multi-attribute steering---e.g., independently controlling style and color palette---would require either multiple bottlenecked codes with disentanglement constraints or a structured latent space.
We leave this extension to future work.

\paragraph{Training data dependency.}
SteeringDiffusion requires a small set of style-exemplar images for training (here, $\sim$250 from ArtBench-10 per style).
This is substantially less data than full fine-tuning but is not zero-shot.
Training-free style transfer methods~\cite{wang2023stylediffusion,meng2022sdedit} are complementary: they trade off controllability for zero data requirements.

\paragraph{UNet-specific architecture.}
Our implementation targets the GroupNorm-based residual blocks of SD~1.5 and SDXL UNets.
Transformer-based diffusion architectures such as DiT~\cite{peebles2023dit} use different normalization schemes (e.g., AdaLN-Zero), requiring architectural adaptation of the steering injector.
The S-BEC principle---bottlenecked activation modulation with temporal gating---is architecture-agnostic, but the specific injection mechanism is not.

\paragraph{Concept erasure and safety.}
Activation steering has been applied for concept erasure and safety alignment in language models~\cite{zou2023representation,gandikota2023erasing}.
Extending S-BEC to \emph{suppress} concepts (negative steering) or enforce safety constraints in diffusion models is a promising but unexplored direction that requires careful evaluation beyond the scope of this paper.

\paragraph{Perceptual evaluation.}
Our metrics (CLIP-I, CLIP-T, Style Shift, LPIPS) are standard but imperfect proxies for human perception.
We do not include human evaluation studies in this work.
While Supplementary Sec.~E.1 validates metric robustness across prototype constructions and CLIP backbones, a formal user study would strengthen claims about perceptual quality at extreme steering scales.

% -----------------------------------------------------------------
\subsection{Reproducibility and Implementation Summary}
\label{sec:reproducibility}

To facilitate reproduction of all results, we summarize the complete experimental pipeline.
Full details including code-level specifics are in Supplementary Sec.~A.

\paragraph{Model and data.}
We use Stable Diffusion 1.5 (SD~1.5, 859.5M UNet parameters, frozen throughout) and SDXL for generalization experiments.
Training data consists of ArtBench-10~\cite{liao2022artbench} style subsets ($\sim$250 images per style).
All evaluation uses a fixed set of 20 diverse English prompts $\times$ multiple seeds, generating 553 prompt--seed pairs in total.

\paragraph{Steering architecture.}
The steering module consists of:
(i)~a prompt encoder $g_\theta$ that maps the CLIP text embedding to a latent code $v \in \mathbb{R}^k$;
(ii)~per-block linear projections $\gamma_\ell, \beta_\ell: \mathbb{R}^k \to \mathbb{R}^{C_\ell}$ producing FiLM-style affine parameters;
(iii)~a global scalar scale $s \in [0, 2]$ and a timestep gate $f(t)$.
All projections are zero-initialized to guarantee this zero-scale equivalence.
The default configuration uses $k{=}16$ and injects into \texttt{mid\_up} blocks (9 blocks total), yielding 0.88M trainable parameters (0.10\% of UNet).

\paragraph{Timestep gate.}
The gate $f(t) \in [0,1]$ is a shifted-scaled sigmoid that suppresses steering during early denoising steps (where global structure forms) and ramps modulation toward later steps (where texture and style are refined).
Supplementary Sec.~E.2 ablates gating versus uniform application, showing that temporal gating is critical for maintaining content preservation at high steering scales.

\paragraph{Training protocol.}
AdamW optimizer, learning rate $5 \times 10^{-5}$, batch size 4, cosine schedule with gradient clipping.
All configurations train for 5{,}000 steps ($\sim$20 minutes on a single NVIDIA A6000).
The diffusion loss (standard $\epsilon$-prediction MSE) is used without auxiliary losses.

\paragraph{Inference and evaluation.}
Generation uses DDIM~\cite{song2021ddim} with $T{=}30$ steps and classifier-free guidance scale 7.5.
For inversion-stability experiments, we use guidance scale 1.0 (no CFG) during inversion and decode with VAE posterior mean (no sampling noise).
All CLIP metrics use ViT-L/14~\cite{clip} unless otherwise noted; Supplementary Sec.~C.1 confirms robustness under ViT-B/32.

\paragraph{Baseline training.}
LoRA baselines use identical data, training budget, and optimizer.
LoRA rank $r{=}4$ (0.80M params) matches the steering parameter budget; $r{=}8$ (1.59M params) serves as a higher-capacity comparison.
ControlNet uses the official architecture with 361M trainable parameters.
All baselines are trained on the same Art Nouveau subset with the same number of steps for fair comparison.

\paragraph{Ablation summary}
We summarize the most essential ablation findings here (full curves in Supplementary).
(i) Temporal gating is critical for monotonic control; removing it introduces content disruption and non-monotonicity at high scales.
(ii) Injection is robust within mid-to-up layers; uniform injection degrades content without improving stylization.
(iii) Control dimensionality shows stable monotonic trade-offs across $k\in\{4,8,16\}$; larger $k$ mainly affects peak stylization.
(iv) Even with rank-stabilized scaling and matched placement, LoRA exhibits less consistent trade-offs at high multipliers.

% =====================================================================
% 4. CONCLUSION
% =====================================================================
\section{Conclusion}
\label{sec:conclusion}

We presented SteeringDiffusion, a lightweight activation-level control interface for diffusion models that exposes a smooth, monotonic, and style-invariant control surface.
Through the lens of Steering via Bottlenecked Explicit Control (S-BEC), our method demonstrates that frozen diffusion backbones can be effectively steered by a single scalar at inference time without retraining.

Our systematic comparison shows that existing methods---LoRA (weight-space), ControlNet (external branch), and rank-1 adapters (fixed direction)---cannot expose a comparable control surface.
LoRA exhibits style collapse at high multipliers, ControlNet requires orders of magnitude more parameters with severe content disruption, and rank-1 adapters lack the condition-awareness needed for smooth control.
In contrast, SteeringDiffusion maintains strict monotonicity across four diverse artistic styles, generalizes from SD~1.5 to SDXL, and incurs negligible runtime overhead.

We focus on style adaptation as a concrete testbed; extensions to concept erasure, negative guidance, safety-oriented control, and Transformer-based diffusion architectures~\cite{peebles2023dit} are promising directions for future work.

%%%%%%%%%%%%%%%%%%%%%%%%%%%%%%%%%%%%%%%%%%%%%%%%%%%%%%%%%%%%

\bibliographystyle{unsrt}
\bibliography{references}

\end{document}